\documentclass[11pt]{article}
\usepackage{coling2016}
\usepackage{times}
\usepackage{url}
\usepackage{latexsym}

\usepackage{epsfig}
\usepackage{multirow}
\usepackage{graphicx}
\usepackage{amssymb}
\usepackage{amsmath}
\usepackage{amsfonts}
\usepackage{subfigure}
\usepackage{color}

\title{Translation of Patent Sentences with a Large Vocabulary \\of
Technical Terms Using Neural Machine Translation}

\author{
  \begin{tabular}{c}
   Zi Long \\
   Takehito Utsuro
  \end{tabular} \\
  Grad. Sc. Sys. \& Inf. Eng., \\
  University of Tsukuba, \\
  sukuba, 305-8573, Japan \\
  \\
  \And
  Tomoharu Mitsuhashi\\
  Japan Patent \\
  Information Organization, \\
  4-1-7, Tokyo, Koto-ku, \\
  Tokyo, 135-0016, Japan \\
  \\
  \And
  Mikio Yamamoto \\
  Grad. Sc. Sys. \& Inf. Eng., \\
  University of Tsukuba, \\
  Tsukuba, 305-8573, Japan \\
  \\}

\date{}

\begin{document}
\maketitle
\begin{abstract}
 Neural machine translation (NMT), a new approach to machine
 translation, has achieved promising results comparable to those of traditional
 approaches such as statistical machine translation (SMT). Despite its recent success, NMT
 cannot handle a larger vocabulary because training complexity and
 decoding complexity proportionally increase with the number of target words. This problem
 becomes even more serious when translating patent documents, 
 which contain many technical terms that are observed infrequently.
 In NMTs, words that are out of vocabulary are
 represented by a single unknown token.
 In this paper, we propose a method that enables NMT to translate patent sentences comprising a
 large vocabulary of technical terms. We train an NMT system on
 bilingual data wherein technical terms are replaced with
 technical term tokens; this allows it to translate most of the source
 sentences except technical terms. Further, we use it as a decoder to
 translate source sentences with technical term tokens and replace the
 tokens with technical term translations using SMT. We also use it to
 rerank the 1,000-best SMT translations on the basis of the average of the SMT
 score and that of the NMT rescoring of the translated sentences
 with technical term tokens. Our experiments on Japanese-Chinese patent
 sentences show that the proposed NMT system achieves a substantial
 improvement of up to 3.1 BLEU points and 2.3 RIBES points over traditional SMT systems and an
 improvement of approximately 0.6 BLEU points and 0.8 RIBES points over an equivalent
 NMT system without our proposed technique.
\end{abstract}

\section{Introduction}
\label{sec:inro}

\begin{figure*}
   \centering
   \includegraphics[scale=0.53]{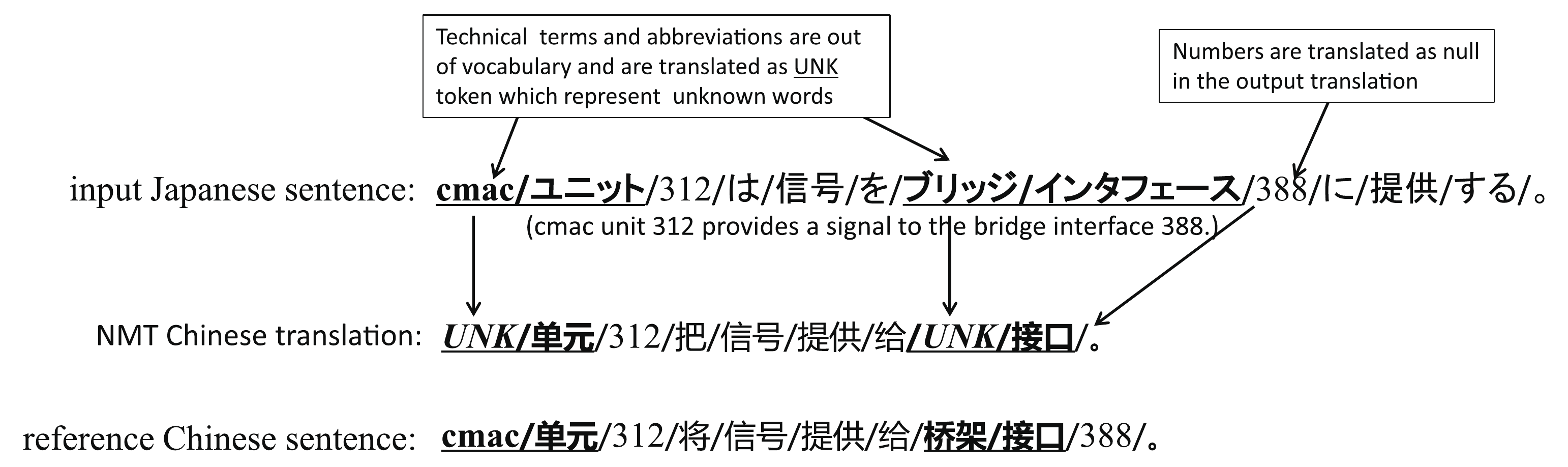}
   \caption{Example of translation errors when translating patent
   sentences with technical terms using NMT}
   \label{fig:problem}
  \end{figure*}

  Neural machine translation (NMT), a new approach to solving machine
  translation, has achieved promising results
  \cite{Kalch13,Sutskever14,Cho14,Bahdanau15,Jean15,Luong15b,Luong15}.
  An NMT system builds a simple large neural network that reads the
  entire input source sentence and generates an output translation. The
  entire neural network is jointly trained to maximize the conditional
  probability of a correct translation of a source sentence with a
  bilingual corpus. Although NMT offers many advantages over traditional
  phrase-based approaches, such as a small memory footprint and simple
  decoder implementation, conventional NMT is limited when it comes to
  larger vocabularies. This is because the training complexity and
  decoding complexity proportionally increase with the number of target
  words. Words that are out of vocabulary are represented by a single
  unknown token in translations, as illustrated in Figure~\ref{fig:problem}.
  The problem becomes more serious when translating patent documents, which contain
  several newly introduced technical terms.
  
  There have been a number of related studies that address the
  vocabulary limitation of NMT systems.
  Jean el al.~\shortcite{Jean15} provided  
  an efficient approximation to the softmax to accommodate a very large
  vocabulary in an NMT system. 
  Luong et al.~\shortcite{Luong15} proposed annotating the occurrences of
  a target unknown word token with positional information to track its
  alignments, after which they replace the tokens
  with their translations using simple word dictionary lookup or
  identity copy. 
  Li et al.~\shortcite{Li16} proposed to replace out-of-vocabulary 
  words with similar in-vocabulary words 
  based on a similarity model learnt from monolingual data.
  Sennrich et al.~\shortcite{Sennrich16} introduced an effective approach based on encoding  
  rare and unknown words as sequences of subword units.
  Luong and Manning~\shortcite{Luong16} provided a
  character-level and word-level hybrid NMT model to achieve an open
  vocabulary, and Costa-juss\`{a} and Fonollosa~\shortcite{Jussa16} proposed 
  a NMT system based on character-based embeddings.
  
  However, these previous approaches have limitations when translating
  patent sentences. This is because
  their methods only focus on addressing the problem of unknown words even
  though the words are
  parts of technical terms. It is obvious that a technical term should be
  considered as one word that comprises 
  components that always have different meanings and translations when
  they are used alone. An
  example is shown in Figure\ref{fig:problem}, wherein Japanese word
  ``\includegraphics[scale=0.45]{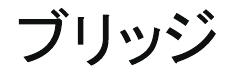}''(bridge) 
  should be translated to Chinese word
  ``\includegraphics[scale=0.45]{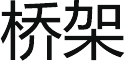}'' 
  when included in technical term ``bridge interface''; however, it is
  always translated as ``\includegraphics[scale=0.45]{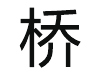}''.
  
  In this paper, we propose a method that enables NMT to translate patent
  sentences with a large vocabulary of technical terms. 
  We use an NMT model similar to that used by Sutskever et
  al.~\shortcite{Sutskever14}, which uses a deep long short-term memories
  (LSTM)~\cite{Hochreiter97} to encode the
  input sentence and a separate deep LSTM to output the translation.
  We train the NMT model on a bilingual corpus in which the technical terms are replaced
  with technical term tokens; this allows it to translate most of the
  source sentences except technical terms.
  Similar to Sutskever et al.~\shortcite{Sutskever14}, 
  we use it as a decoder to translate source sentences with technical term
  tokens and replace the tokens with technical term translations using
  statistical machine translation (SMT). We also use it to rerank the
  1,000-best SMT translations on the basis of the average of the SMT and
  NMT scores of the translated sentences that have been rescored with
  the technical term tokens. 
  Our experiments on Japanese-Chinese patent sentences show that
  our proposed NMT system achieves
  a substantial improvement of up to 3.1 BLEU points and 2.3 RIBES points
  over a traditional SMT
  system and an improvement of approximately 0.6 BLEU points and 0.8 RIBES
  points over an equivalent NMT system without our proposed technique.
 
\section{Japanese-Chinese Patent Documents}
\label{sec:patent}
 
 Japanese-Chinese parallel patent documents were collected from the
 Japanese patent documents published
 by the Japanese Patent Office (JPO) during 2004-2012 and the Chinese
 patent documents published by the State
 Intellectual Property Office of the People's Republic of China (SIPO)
 during 2005-2010. From the collected documents, we
 extracted 312,492 patent families, and the method of 
 Utiyama and Isahara~\shortcite{Uchiyama07bs}
 was applied\footnote{
  Herein, we used a Japanese-Chinese translation lexicon comprising
  around 170,000 Chinese entries.
 } 
 to the text of the extracted patent families to align the Japanese and Chinese
 sentences. 
 The Japanese sentences were segmented into a sequence of morphemes using the
 Japanese morphological analyzer MeCab\footnote{
   \url{http://mecab.sourceforge.net/}
 }
 with the morpheme lexicon IPAdic,\footnote{
   \url{http://sourceforge.jp/projects/ipadic/}
 }
 and the Chinese sentences
 were segmented into a sequence of words using the Chinese
 morphological analyzer Stanford Word
 Segment~\cite{Tseng05a} trained using the Chinese Penn
 Treebank. 
 In this study, Japanese-Chinese parallel patent sentence
 pairs were ordered in descending order of sentence-alignment score
 and we used the topmost 2.8M pairs, 
 whose Japanese sentences contain fewer than 40 morphemes and
 Chinese sentences contain fewer than 40 words.\footnote{
  In this paper, we focus on the task of translating patent sentences
  with a large vocabulary of technical terms using the NMT system, where
  we ignore the translation task of patent sentences that are longer than
  40 morphemes in Japanese side or longer than 40 words in Chinese side.
 }

\section{Neural Machine Translation (NMT)}
\label{sec:nmt}

 NMT uses a single
 neural network trained jointly to maximize the translation performance
 ~\cite{Kalch13,Sutskever14,Cho14,Bahdanau15,Luong15b}. 
 Given a source sentence {\boldmath $x$} $=(x_1,\ldots,x_N)$ and target
 sentence {\boldmath $y$} $=(y_1,\ldots,y_M)$, an NMT system uses a neural network to
 parameterize the conditional distributions
 \begin{eqnarray}
  p(y_l \mid y_{< l},\mbox{\boldmath $x$}) \nonumber
 \end{eqnarray}
 for $1 \leq l \leq M$. Consequently, it becomes possible to compute and
 maximize the log probability of the target sentence given the source
 sentence
 \begin{eqnarray}
  \label{eq:pro}
  \log p(\mbox{\boldmath $y$} \mid \mbox{\boldmath $x$}) = \sum_{l=1}^{M} \log p(y_l|y_{< l},\mbox{\boldmath $x$})
 \end{eqnarray}

 In this paper, we use an NMT model similar to that used by Sutskever et
 al.~\shortcite{Sutskever14}. It uses two separate deep LSTMs
 to encode the input sequence and output the translation. The encoder,
 which is implemented as a recurrent neural network, reads the source sentence
 one word at a time and then encodes it into a large vector that represents the entire source
 sentence. The decoder, another recurrent neural network, generates a
 translation on the basis of the encoded vector one word at a time.

 One important difference between our NMT model and the one used by
 Sutskever et al.~\shortcite{Sutskever14} is
 that we added an attention mechanism. Recently, Bahdanau et al.~\shortcite{Bahdanau15}
 proposed an attention mechanism, a form of random access memory, to help NMT cope with long input
 sequences. Luong et al.~~\shortcite{Luong15b} proposed an attention mechanism for
 different scoring functions in order to compare the source and target
 hidden states as well as different strategies for placing the attention. In
 this paper, we utilize the attention mechanism proposed by Bahdanau et
 al.~\shortcite{Bahdanau15}, wherein each output target word is predicted on the basis
 of not only a recurrent
 hidden state and the previously predicted word but also a context vector
 computed as the weighted sum of the hidden states.
  
\section{NMT with a Large Technical Term Vocabulary}
\label{sec:trans}

  \begin{figure*}
   \centering
   \includegraphics[scale=0.38]{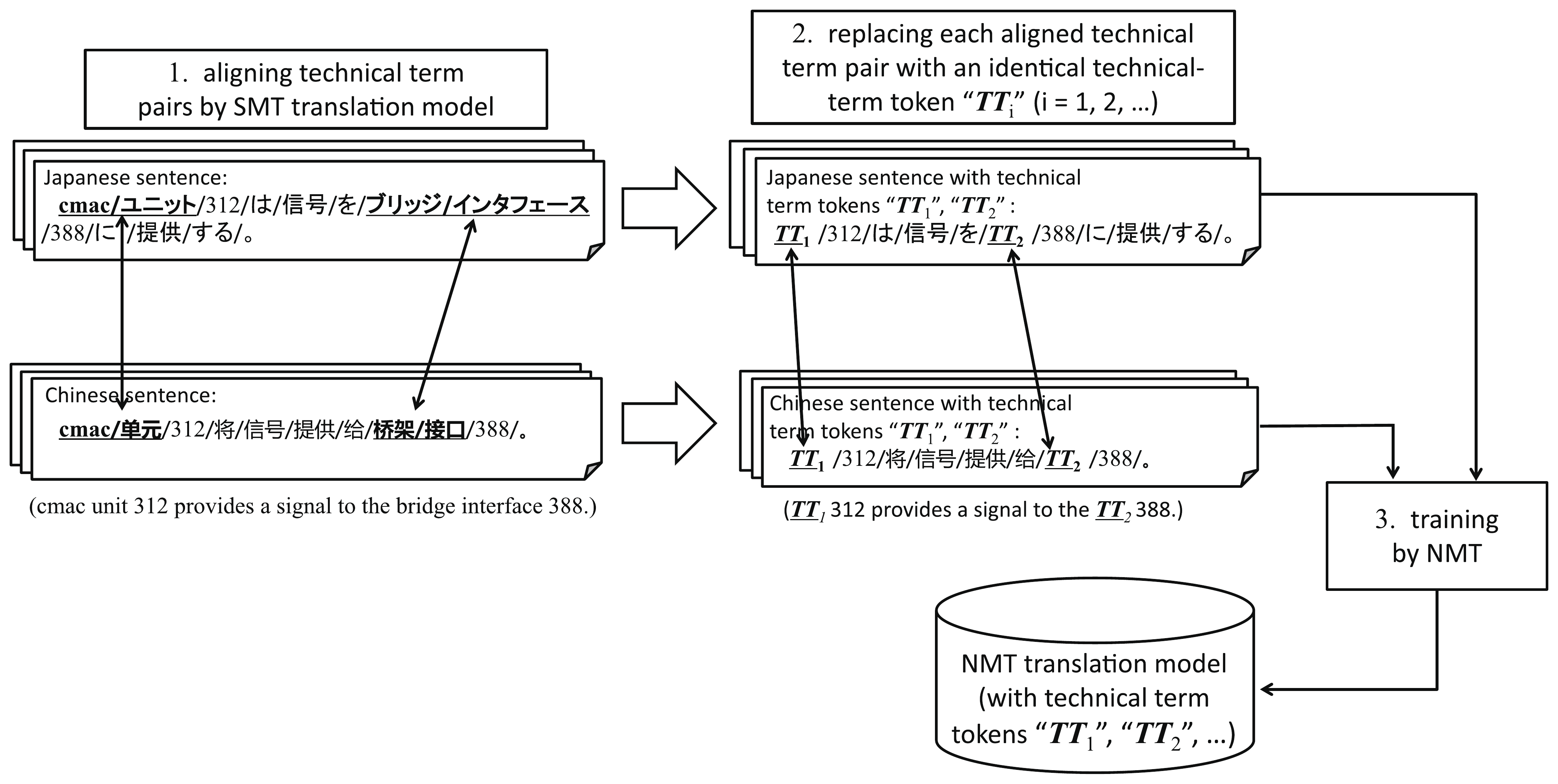}
   \caption{NMT training after replacing technical term pairs with
   technical term tokens ``$TT_{i}$'' ($i=1,2,\ldots$)}
   \label{fig:train}
  \end{figure*}

\subsection{NMT Training after Replacing Technical Term Pairs with Tokens}
\label{subsec:train}

  Figure~\ref{fig:train} illustrates the procedure of the
  training model with parallel
  patent sentence pairs, wherein technical terms are replaced with
  technical term tokens ``$TT_{1}$'', ``$TT_{2}$'', $\ldots$.

  In the step 1 of Figure~\ref{fig:train}, 
  we align the Japanese technical terms, which are automatically
  extracted from the Japanese sentences,
  with their Chinese translations in the Chinese sentences.\footnote{
   In this work, we approximately regard all the Japanese compound nouns 
   as Japanese technical terms. 
   These Japanese compound nouns are automatically 
   extracted by simply concatenating a
   sequence of morphemes whose parts of speech are either nouns,
   prefixes, suffixes, unknown words, numbers, or alphabetical
   characters. Here, morpheme sequences starting or ending with certain
   prefixes are inappropriate as Japanese technical terms and are
   excluded. The sequences that include symbols or numbers are also
   excluded. 
   In Chinese side, on the other hand, 
   we regard Chinese translations of extracted Japanese
   compound nouns as Chinese technical terms, where we do not regard
   other Chinese phrases as technical terms.
  }
  Here, we introduce the following two steps to 
  identify technical term pairs in the bilingual Japanese-Chinese corpus:
  \begin{enumerate}
   \item\label{item:pt} According to the approach proposed by Dong et
	al.~\shortcite{Dong15b}, we identify Japanese-Chinese technical term pairs using an SMT
	phrase translation table. Given a parallel sentence pair
	$\langle S_J, S_C\rangle$ containing a Japanese technical term $t_J$, 
	the Chinese translation candidates collected from the phrase translation
	table are matched against the Chinese sentence $S_C$ of the parallel
	sentence pair. Of those found in $S_C$, $t_C$ with the largest translation probability
	$P(t_C\mid t_J)$ is selected, and the bilingual technical term pair $\langle t_J,t_C\rangle$ is
	identified.
   \item For the Japanese technical terms whose Chinese translations are
	 not included in the results of Step~\ref{item:pt}, we then
	 use an approach based on SMT word alignment. Given a parallel sentence
	 pair $\langle S_J, S_C\rangle$ containing
	 a Japanese technical term  $t_J$, a sequence of Chinese words is selected
	 using SMT word alignment,
	 and we use the Chinese translation $t_C$ for the Japanese technical term
	 $t_J$.\footnote{
	  We discard discontinuous sequences and only use
	  continuous ones.
	 }
  \end{enumerate} 

  \begin{figure*}
   \centering
   \includegraphics[scale=0.38]{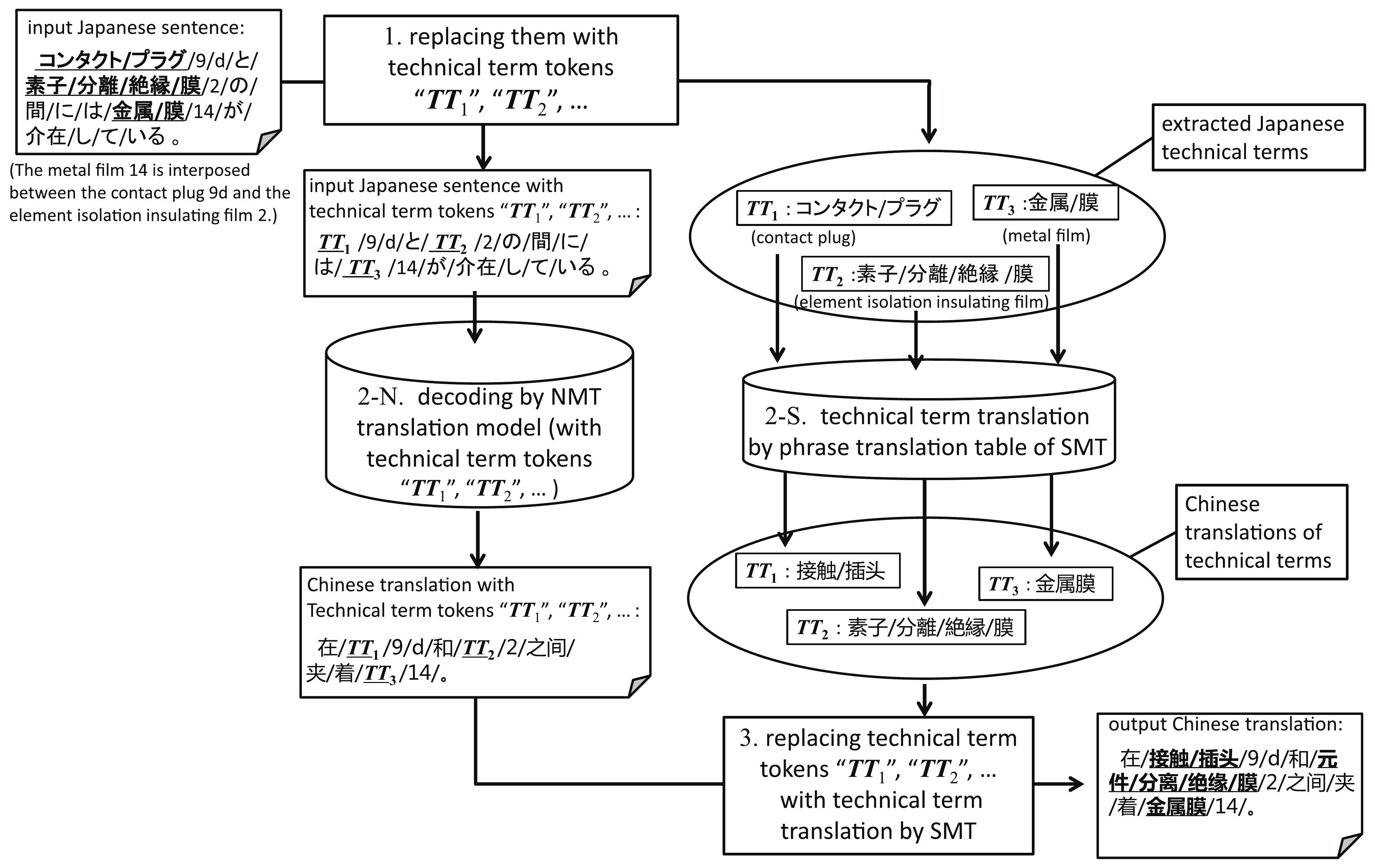}
     \caption{NMT decoding with technical term tokens ``$TT_{i}$''
   ($i=1,2,\ldots$) and SMT technical term translation}
   \label{fig:decode}
  \end{figure*}

  As shown in the step 2 of Figure~\ref{fig:train}, 
  in each of Japanese-Chinese parallel patent sentence pairs, occurrences
  of technical term pairs $\langle t_J^{\ 1},t_C^1 \rangle$, $\langle
  t_J^2,t_C^2\rangle$, $\ldots$, $\langle t_J^k,t_C^k\rangle$ are then
  replaced with technical term tokens $\langle TT_{1},TT_{1} \rangle$,
  $\langle TT_{2},TT_{2} \rangle$, $\ldots$,
  $\langle TT_{k},TT_{k} \rangle$. Technical term pairs $\langle t_J^{1},t_C^1 \rangle$,
  $\langle t_J^2,t_C^2\rangle$, $\ldots$, $\langle t_J^k,t_C^k\rangle$ 
  are numbered in the order of occurrence of Japanese technical terms
  $t_J^{\ i}$ ($i=1,2,\ldots,k$) in each Japanese sentence $S_J$. 
  Here, note that in
  all the parallel sentence pairs $\langle S_J, S_C\rangle$, 
  technical term tokens ``$TT_{1}$'', ``$TT_{2}$'', $\ldots$ that are
  identical throughout all the parallel sentence pairs are used in this procedure.
  Therefore, for example, in all the Japanese patent sentences $S_J$, the Japanese technical
  term $t_J^{\ 1}$ which appears earlier than other Japanese technical terms
  in $S_J$ is replaced with $TT_{1}$.
  We then train the NMT system on a bilingual
  corpus, in which the technical term pairs is
  replaced by ``$TT_{i}$'' ($i=1,2,\ldots$) tokens, 
  and obtain an NMT model in which the technical
  terms are represented as technical term tokens.\footnote{
   We treat the NMT system as a black box, and the strategy we present
   in this paper could be applied to any NMT system
   ~\cite{Kalch13,Sutskever14,Cho14,Bahdanau15,Luong15b}.
  }

\subsection{NMT Decoding and SMT Technical Term Translation} 
\label{subsec:decode}

  Figure~\ref{fig:decode} illustrates the procedure for 
  producing Chinese translations via decoding the Japanese sentence 
  using the method proposed in this paper.
  In the step 1 of Figure~\ref{fig:decode}, 
  when given an input Japanese sentence, 
  we first automatically extract the technical terms and
  replace them with the technical term tokens ``$TT_{i}$''
  ($i=1,2,\ldots$). Consequently, we have an input sentence in which the
  technical term tokens ``$TT_{i}$'' ($i=1,2,\ldots$) represent the
  positions of the technical terms and a list of
  extracted Japanese technical terms. 
  Next, as shown in the step 2-N of Figure~\ref{fig:decode}, 
  the source Japanese sentence with technical term tokens is translated
  using the NMT model trained according to the procedure described in
  Section~\ref{subsec:train},
  whereas the extracted Japanese technical terms are translated using an
  SMT phrase translation table in the step 2-S of
  Figure~\ref{fig:decode}.\footnote{
   We use the translation with the highest probability in the phrase
   translation table. When an input 
   Japanese technical term has multiple translations with the same highest
   probability or has no translation in the phrase translation table, we
   apply a compositional translation generation approach, wherein
   Chinese translation is generated compositionally from the constituents
   of Japanese technical terms.
  }
  Finally, in the step 3, 
  we replace the technical term tokens ``$TT_{i}$'' ($i=1,2,\ldots$) of
  the sentence translation with SMT the technical term translations.

\subsection{NMT Rescoring of 1,000-best SMT Translations}
\label{subsec:rescore}

  \begin{figure*}
   \centering
   \includegraphics[scale=0.34]{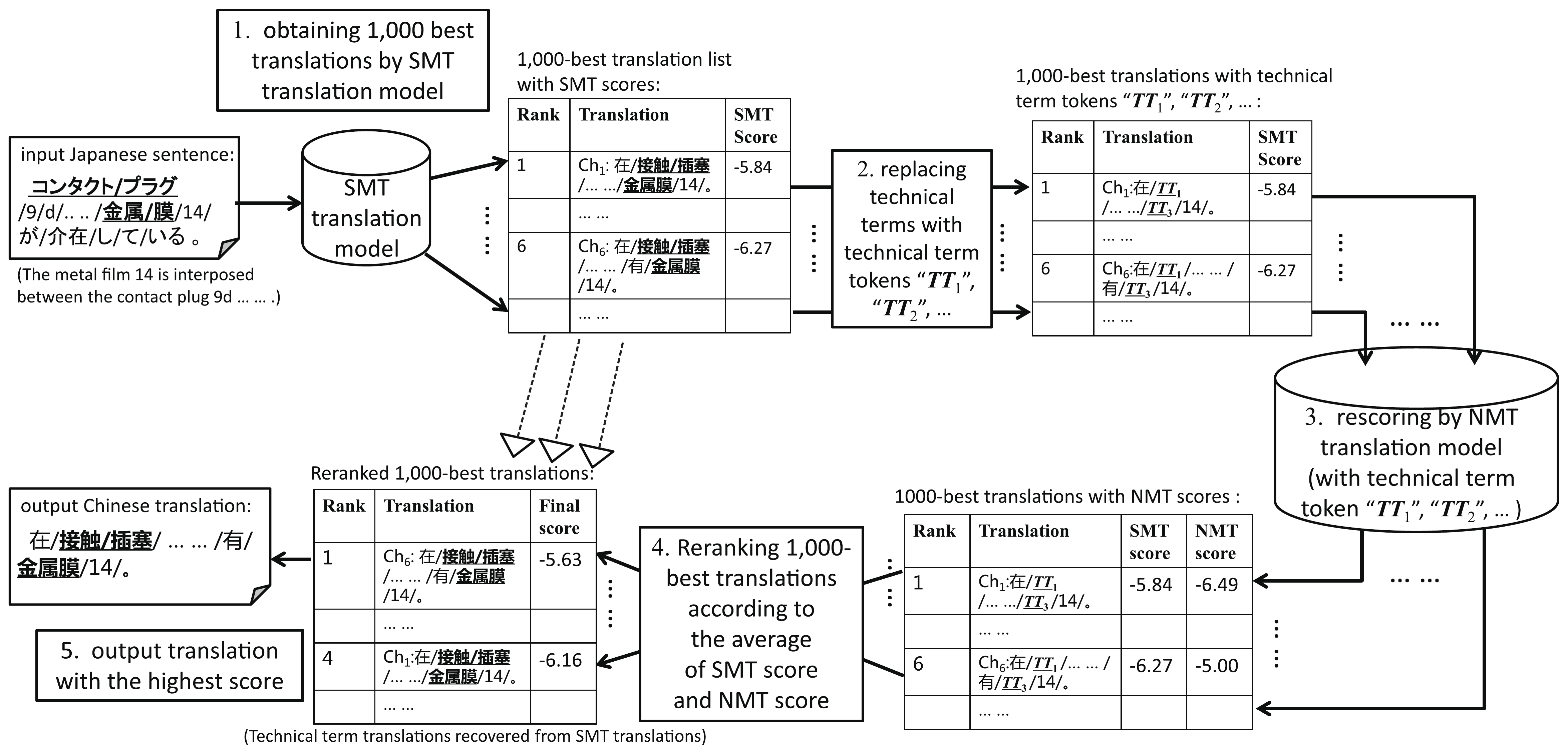}
     \caption{NMT rescoring of 1,000-best SMT translations with
   technical term tokens ``$TT_{i}$''  ($i=1,2,\ldots$)}
   \label{fig:rescore}
  \end{figure*}

  As shown in the step 1 of Figure~\ref{fig:rescore}, 
  similar to the approach of NMT rescoring provided 
  in Sutskever et al.\shortcite{Sutskever14}, 
  we first obtain 1,000-best translation list of the given Japanese
  sentence using the SMT system.
  Next, in the step 2, we then replace the technical 
  terms in the translation sentences with
  technical term tokens ``$TT_{i}$'' ($i = 1,2,3,\ldots$),
  which must be the same with the tokens of their source Japanese technical
  terms in the input Japanese sentence. The
  technique used for aligning Japanese technical terms with their Chinese
  translations is the same as that described in
  Section~\ref{subsec:train}. 
  In the step 3 of Figure~\ref{fig:rescore}, 
  the 1,000-best translations, in which
  technical terms are represented as tokens, are rescored using the NMT
  model trained according to the procedure described in Section~\ref{subsec:train}. 
  Given a Japanese sentence $S_J$ and its 1,000-best Chinese translations
  $S_C^{\ n}$ ($n=1,2,\ldots,\ 1,000$) translated by the SMT system, NMT score of each 
  translation sentence pair $\langle S_J, S_C^n \rangle$ is computed as the log
  probability $\log p(S_C^n \mid S_J)$ of Equation (\ref{eq:pro}).
  Finally, we rerank the 1,000-best translation list on the basis of 
  the average SMT and NMT scores and output the translation with the highest final score.

\section{Evaluation}
\label{sec:exp}

\subsection{Training and Test Sets}
\label{subsec:data}

  We evaluated the effectiveness of the proposed NMT system in translating
  the Japanese-Chinese parallel patent
  sentences described in Section~\ref{sec:patent}. 
  Among the 2.8M parallel sentence
  pairs, we randomly extracted 1,000 sentence pairs for the test set and
  1,000 sentence pairs for the development set; the remaining sentence
  pairs were used for the training set.

  According to the procedure of Section~\ref{subsec:train},
  from the Japanese-Chinese sentence pairs of the training set, 
  we collected 6.5M occurrences of technical term pairs, which are 1.3M
  types of technical term pairs with 800K unique types of
  Japanese technical terms and 1.0M unique types of 
  Chinese technical terms.
  Out of the total 6.5M occurrences of technical term pairs,
  6.2M were replaced with technical term tokens using
  the phrase translation table, while the remaining 300K were replaced
  with technical term tokens using the word alignment.\footnote{
   There are also Japanese technical terms (3\% of all the extracted terms)
   for which Chinese translations can be identified using neither the
   SMT phrase translation table nor the SMT word alignment.
  }
  We limited both the Japanese vocabulary (the source language) and
  the Chinese vocabulary (the target language) to 40K most frequently
  used words.

  Within the total 1,000 Japanese patent sentences in the test set,
  2,244 occurrences of Japanese technical terms were identified, 
  which correspond to 1,857 types.

\subsection{Training Details}
\label{subsec:detail}
 
  For the training of the SMT model, 
  including the word alignment and the phrase
  translation table, we used Moses~\cite{Koehn07as}, a toolkit for a
  phrase-based SMT models.

  For the training of the NMT model, 
  our training procedure and hyperparameter choices were similar to those
  of Sutskever et al.~\shortcite{Sutskever14}.
  We used a deep LSTM neural network
  comprising three layers, with 512 cells in each layer, and a 512-dimensional word
  embedding. Similar to Sutskever et al. (2014), we
  reversed the words in the source sentences
  and ensure that all sentences in a minibatch are roughly the same
  length.
  Further training details are given below:
  \begin{itemize}
   \item All of the LSTM's parameter were initialized with a uniform
	 distribution ranging between -0.06 and 0.06.
   \item We set the size of a minibatch to 128.
   \item We used the stochastic gradient descent, beginning at a
	 learning rate of 0.5. We computed the perplexity of
	 the development set using the currently produced NMT model after
	 every 1,500 minibatches were trained and multiplied the learning rate by 0.99 when
	 the perplexity did not decrease with respect to the last three
	 perplexities. We trained our model for a total of 10 epoches.
   \item Similar to Sutskever et al.~\shortcite{Sutskever14}, we rescaled the
	 normalized gradient to ensure that its norm does not exceed 5.
  \end{itemize}
  We implement the NMT system using TensorFlow,\footnote{
   \url{https://www.tensorflow.org/}
  }
  an open source library for numerical computation.
  The training time was around two days when using the described
  parameters on an 1-GPU machine.

  \begin{table*}
   \begin{center}
    \caption{Automatic evaluation results}
    \label{tab:eva_result}
    \begin{tabular}{|l||c|c|c|c|}
     \hline
     \multirow{2}{*}{System} 
     & \multicolumn{2}{|c|}{
     \begin{tabular}{c}
      NMT decoding and \\
      SMT technical term \\
      translation
     \end{tabular}}
     & \multicolumn{2}{|c|}{
	 \begin{tabular}{c}
	  NMT rescoring of\\
	  1,000-best SMT\\
	  translations
	 \end{tabular}}
     \\ \cline{2-5}
     
     & BLEU & RIBES & BLEU & RIBES 
     \\ \hline \hline

     Baseline SMT~\cite{Koehn07as} & 52.5 & 88.5 & - & -
     \\ \hline
     
     Baseline NMT & 53.5 & 90.0 & 55.0 & 89.1 
     \\ \hline

     NMT with technical term translation by SMT 
     & 55.3 & {\bf 90.8} & {\bf 55.6} & 89.3
     \\ \hline
    \end{tabular}
   \end{center}
  \end{table*}

  \begin{table*}
   \begin{center}
    \caption{Human evaluation results (the score of pairwise evaluation 
    ranges from $-$100 to 100 and the score of JPO adequacy evaluation ranges from 1 to 5)}
    \label{tab:he_result}
    \begin{tabular}{|l||c|c|c|c|}
     \hline
     \multirow{3}{*}{System} 
     & \multicolumn{2}{|c|}{
     \begin{tabular}{c}
      NMT decoding and \\
      SMT technical term \\
      translation
     \end{tabular}}
     & \multicolumn{2}{|c|}{
         \begin{tabular}{c}
          NMT rescoring of\\
          1,000-best SMT\\
          translations
         \end{tabular}}
     \\ \cline{2-5}

     & $\!\!\!$\begin{tabular}{c}
	pairwise\\evaluation
       \end{tabular}$\!\!\!$
      & $\!\!\!$\begin{tabular}{c}
	 JPO\\adequacy\\evaluation
	\end{tabular}$\!\!\!$
      & $\!\!\!$\begin{tabular}{c}
        pairwise\\evaluation
       \end{tabular}$\!\!\!$
      & $\!\!\!$\begin{tabular}{c}
         JPO\\adequacy\\evaluation
        \end{tabular}$\!\!\!$
     \\ \hline \hline

     Baseline SMT~\cite{Koehn07as} & - & 3.5 & - & -
     \\ \hline\hline
     
     Baseline NMT & 5.0 & 3.8 & 28.5 & 4.1 
     \\ \hline \hline

     NMT with technical term translation by SMT
     & {\bf 36.5} & {\bf 4.3} & 31.0 & 4.1
     \\ \hline
    \end{tabular}
   \end{center}
  \end{table*}
  
\subsection{Evaluation Results}
\label{subsec:eva}

  We calculated automatic evaluation scores for the translation results
  using two popular metrics:
  BLEU~\cite{papineni-EtAl:2002:ACL} and RIBES~\cite{Isozaki10}. As shown
  in Table~\ref{tab:eva_result}, we report the
  evaluation scores, on the basis of the translations by
  Moses~\cite{Koehn07as}, as the baseline SMT\footnote{
    We train the SMT system on the same training set and tune it with
  development set.
  } and the scores based on translations produced by the equivalent NMT system without our
  proposed approach as the baseline NMT. 
  As shown in Table~\ref{tab:eva_result}, 
  the two versions of the proposed NMT systems 
  clearly improve the translation quality when compared with the baselines.
  When compared with the baseline SMT, the 
  performance gain of the proposed system is approximately 3.1 BLEU
  points if translations are produced by the proposed NMT system
  of Section~\ref{subsec:rescore}
  or 2.3 RIBES points if translations are produced by the proposed NMT
  system of Section~\ref{subsec:decode}. 
  When compared with the result of decoding with the baseline NMT, 
  the proposed NMT system of Section~\ref{subsec:decode} 
  achieved performance gains of 0.8 RIBES points.
  When compared with the result of reranking with the baseline NMT, 
  the proposed NMT system of Section~\ref{subsec:rescore}
  can still achieve performance gains of 0.6 BLEU points.
  Moreover, when the output translations produced by NMT
  decoding and SMT technical term translation described in
  Section~\ref{subsec:decode} with the output translations
  produced by decoding with the baseline NMT, 
  the number of unknown tokens included in output translations reduced
  from 191 to 92.
  About 90\% of remaining unknown tokens correspond to
  numbers, English words, abbreviations, and symbols.\footnote{
   In addition to the two versions of the proposed NMT systems
   presented in Section~\ref{sec:trans}, 
   we evaluated a modified version of the propsed NMT system, where we
   introduce another type of token corresponding to unknown compound nouns and
   integrate this type of token with the technical term token in the procedure
   of training the NMT model. We achieved a slightly improved
   translation performance, BLEU/RIBES scores of 55.6/90.9 for the
   proposed NMT system of Section~\ref{subsec:decode} and those of
   55.7/89.5 for the proposed NMT system of Section~\ref{subsec:rescore}.
   }

  \begin{figure*}
   \centering
   \includegraphics[scale=0.45]{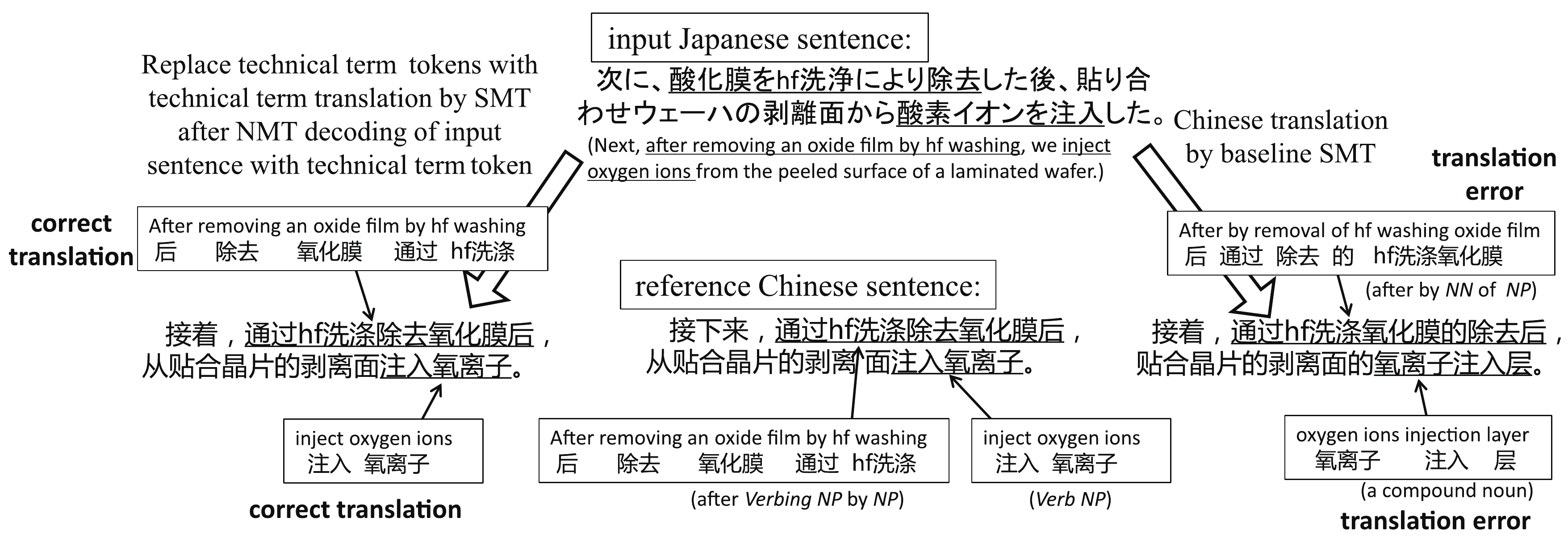}
   \caption{Example of correct translations produced by the proposed NMT system
   with SMT technical term translation (compared with baseline SMT) }
   \label{fig:eva_1}
  \end{figure*}

  In this study, we also conducted two types of human evaluation according to
  the work of Nakazawa et al.~\shortcite{Nakazawa15}: 
  pairwise evaluation and JPO adequacy evaluation.
  During the procedure of pairwise evaluation, we compare
  each of translations produced by the baseline SMT with that produced by
  the two versions of the proposed NMT systems, 
  and judge which translation is better, or whether they are with comparable quality. 
  The score of pairwise evaluation
  is defined by the following formula, where $W$ is the number of
  better translations compared to the baseline
  SMT, $L$ the number of worse translations compared to the baseline
  SMT, and $T$ the number of translations having their quality
  comparable to those produced by the baseline SMT:
  \begin{eqnarray}
   score=100 \times \frac{W-L}{W+L+T} \nonumber
  \end{eqnarray}
  The score of pairwise evaluation ranges from $-$100 to 100. 
  In the JPO adequacy evaluation, Chinese translations are evaluated according to the
  quality evaluation criterion for translated patent documents 
  proposed by the Japanese Patent Office (JPO).\footnote{
   \url{https://www.jpo.go.jp/shiryou/toushin/chousa/pdf/tokkyohonyaku_hyouka/01.pdf}
  (in Japanese)
  }
  The JPO adequacy criterion judges whether or not the technical factors and their relationships
  included in Japanese patent sentences are
  correctly translated into Chinese, and score Chinese
  translations on the basis
  of the percentage of correctly translated information, where the score
  of 5 means all of those information are translated correctly, while that of 1 means most 
  of those information are not translated correctly. 
  The score of the JPO adequacy evaluation is defined as the average over
  the whole test sentences.
  Unlike the study conducted Nakazawa et al.~\cite{Nakazawa15}, 
  we randomly selected 200 sentence pairs from the test set for
  human evaluation, and both human evaluations were conducted using only
  one judgement. Table~\ref{tab:he_result} shows the results of the human evaluation
  for the baseline SMT, the baseline NMT, and the proposed NMT system. We observed
  that the proposed system achieved the best performance for both pairwise evaluation and JPO
  adequacy evaluation when we replaced technical term tokens with SMT
  technical term translations after
  decoding the source sentence with technical term tokens.

  \begin{figure*}
   \centering
   \includegraphics[scale=0.45]{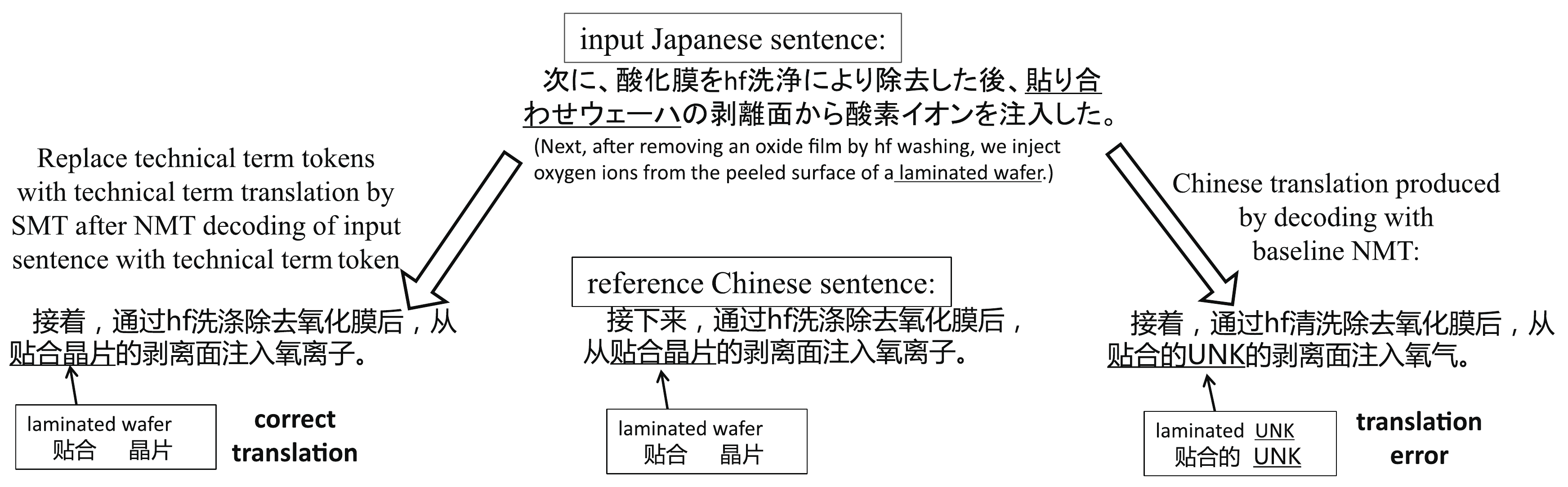}
   \caption{Example of correct translations produced by the proposed NMT system with
   SMT technical term translation (compared to decoding with the baseline NMT)}
   \label{fig:eva_2}
  \end{figure*}

\begin{figure*}
   \centering
   \includegraphics[scale=0.41]{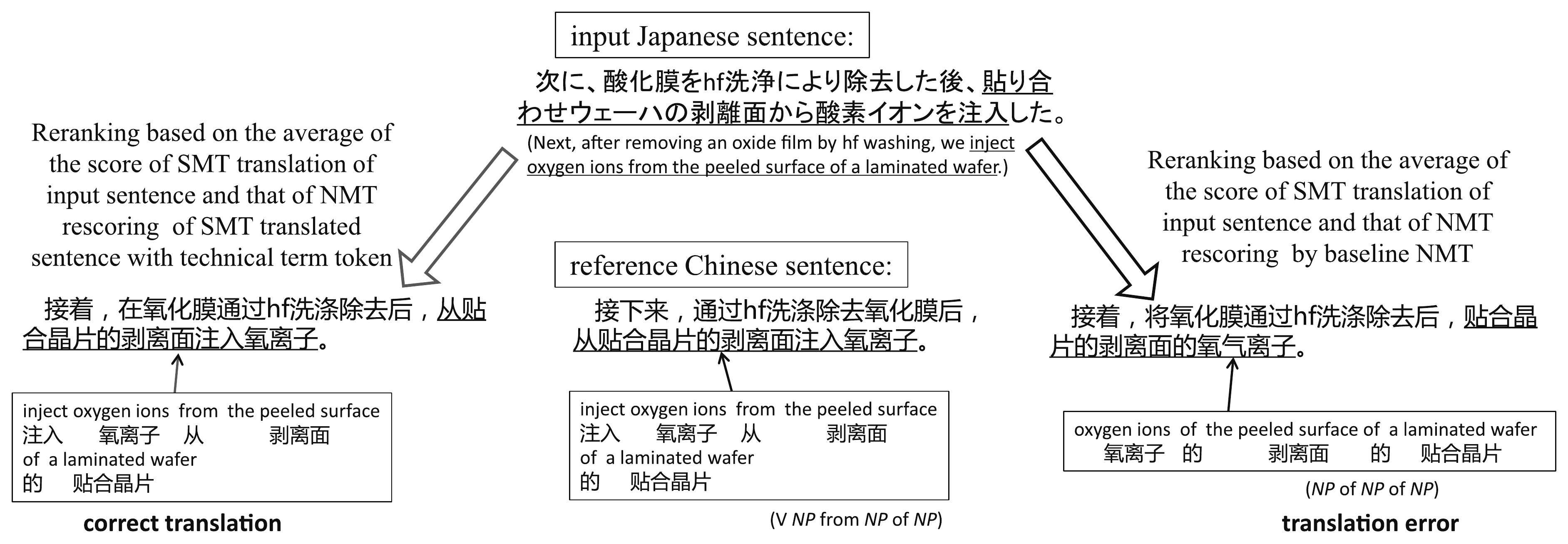}
   \caption{Example of correct translations produced by reranking the 1,000-best
   SMT translations with the proposed NMT system (compared to reranking
   with the baseline NMT)}
   \label{fig:eva_3}
  \end{figure*}

  Throughout Figure~\ref{fig:eva_1}$\sim$Figure~\ref{fig:eva_3}, 
  we show an identical source Japanese sentence and  
  each of its translations produced by the two versions of the proposed NMT
  systems, compared with translations produced by the three
  baselines, respectively.
  Figure~\ref{fig:eva_1} shows an example of correct translation produced by the
  proposed system in comparison to that produced by the baseline
  SMT. In this example, our model correctly translates the Japanese
  sentence into Chinese, whereas the translation by the baseline SMT is a
  translation error with several erroneous syntactic structures.
  As shown in Figure~\ref{fig:eva_2}, the second
  example highlights that the proposed NMT system of
  Section~\ref{subsec:decode} can correctly translate the
  Japanese technical 
  term ``\includegraphics[scale=0.45]{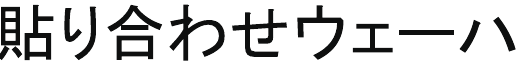}''(laminated
  wafer) to the Chinese technical term
  ``\includegraphics[scale=0.45]{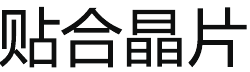}''.
  The translation by the baseline NMT is a translation
  error because of not only the erroneously translated unknown token but
  also the Chinese word
  ``\includegraphics[scale=0.45]{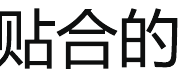}'', 
  which is not appropriate as a component of a Chinese technical term. 
  Another example is shown in Figure~\ref{fig:eva_3}, where
  we compare the translation of a reranking SMT 1,000-best translation
  produced by the proposed NMT system with that produced by reranking
  with the baseline NMT. It
  is interesting to observe that compared with the baseline NMT, we obtain a
  better translation when we rerank the 
  1,000-best SMT translations using the proposed NMT system, in which
  technical term tokens represent technical terms. 
  It is mainly because the correct Chinese translation
  ``\includegraphics[scale=0.45]{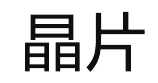}''(wafter) 
  of Japanese word ``\includegraphics[scale=0.45]{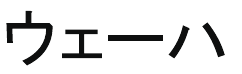}''
  is out of the 40K NMT vocabulary (Chinese), causing reranking with 
  the baseline NMT to produce the translation with an erroneous
  construction of ``noun phrase of noun phrase of noun phrase''. 
  As shown in Figure~\ref{fig:eva_3}, the proposed NMT system of
  Section~\ref{subsec:rescore} produced the translation with a correct
  construction, mainly because Chinese word
  ``\includegraphics[scale=0.45]{figures/wafer_ch.pdf}''(wafter) is
  a part of Chinese technical term
  ``\includegraphics[scale=0.45]{figures/lwafer_ch.pdf}''(laminated
  wafter) and is replaced with a technical term token and then rescored
  by the NMT model (with technical term tokens ``$TT_{1}$'', ``$TT_{2}$'', $\ldots$).

\section{Conclusion}
\label{sec:fin}

  In this paper, we proposed an NMT method capable of translating patent
  sentences with a large vocabulary
  of technical terms. We trained an NMT system on a bilingual corpus,
  wherein technical terms are
  replaced with technical term tokens; this allows it to translate most of
  the source sentences except
  the technical terms. Similar to Sutskever et
  al.~\shortcite{Sutskever14}, we used it as
  a decoder to translate the source sentences with technical term tokens
  and replace the tokens with technical terms
  translated using SMT. We also used it to rerank the 1,000-best SMT translations
  on the basis of the 
  average of the SMT score and that of NMT rescoring of translated
  sentences with technical term tokens. For
  the translation of Japanese patent sentences, we observed that our
  proposed NMT system performs
  better than the phrase-based SMT system as well as the equivalent NMT
  system without our proposed approach.

  One of our important future works is to evaluate our
  proposed method in the NMT system proposed by Bahdanau et
  al.~\shortcite{Bahdanau15}, which introduced a bidirectional recurrent
  neural network as encoder and is the state-of-the-art of pure NMT system
  recently. However, the NMT system proposed by Bahdanau et
  al.~\shortcite{Bahdanau15} also has a limitation in addressing
  out-of-vocabulary words. 
  Our proposed NMT system is expected to improve the translation performance
  of patent sentences by applying approach of Bahdanau et al.~\shortcite{Bahdanau15}.
  Another important future work is to quantitatively 
  compare our study with the work of Luong et
  al.~\shortcite{Luong15}. In the work of Luong et
  al.~\shortcite{Luong15}, they replace low frequency
  single words and translate them in a post-processing Step using a
  dictionary, while we propose to replace the whole technical terms and
  post-translate them with phrase translation table of SMT
  system. Therefore, our proposed NMT system is expected to be appropriate
  to translate patent documents which contain many technical terms 
  comprised of multiple words and should be translated together.
  We will also evaluate the present study by reranking the
  n-best translations produced by the proposed NMT system on the basis
  of their SMT rescoring. Next, we will rerank translations from both the 
  n-best SMT translations and n-best NMT translations. 
  As shown in Section~\ref{subsec:eva}, the decoding approach of our
  proposed NMT system achieved the best RIBES performance and human
  evaluation scores in our experiments, whereas the reranking approach achieved the best
  performance with respect to BLEU. 
  A translation with the
  highest average SMT and NMT scores of the n-best translations produced
  by NMT and SMT, respectively, is expected to be an effective
  translation.


\bibliographystyle{acl}
\bibliography{myabbrv,mydb,long}

\end{document}